\documentclass[letterpaper, 10 pt, conference]{ieeeconf}

\usepackage{graphicx}
\usepackage{balance}
\usepackage{comment}
\usepackage{cite}
\usepackage{amssymb}
\usepackage[tight,footnotesize]{subfigure}
\usepackage[active]{srcltx}
\usepackage{amsmath}

\graphicspath{{./figs/}}
\renewcommand{\baselinestretch}{0.95}
\usepackage{eurosym}
\usepackage[plain]{algorithm}
\usepackage{algorithmic}
\usepackage{multicol}
\usepackage{dsfont}
\usepackage{amsfonts}

\usepackage{cases}
\usepackage{xcolor}

\IEEEoverridecommandlockouts
\begin{document}

\title{UAV-based Receding Horizon Control for 3D Inspection Planning}

\author{Savvas~Papaioannou,~Panayiotis~Kolios,~Theocharis~Theocharides,\\~Christos~G.~Panayiotou~ and ~Marios~M.~Polycarpou
\thanks{The authors are with the KIOS Research and Innovation Centre of Excellence (KIOS CoE) and the Department of Electrical and Computer Engineering, University of Cyprus, Nicosia, 1678, Cyprus. {\tt\small \{papaioannou.savvas, pkolios, ttheocharides, christosp, mpolycar\}@ucy.ac.cy}}
}

\maketitle

\begin{abstract}
Nowadays, unmanned aerial vehicles or UAVs are being used for a wide range of tasks, including infrastructure inspection, automated monitoring and coverage. This paper investigates the problem of 3D inspection planning with an autonomous UAV agent which is subject to dynamical and sensing constraints. We propose a receding horizon 3D inspection planning control approach for generating optimal trajectories which enable an autonomous UAV agent to inspect a finite number of feature-points scattered on the surface of a cuboid-like structure of interest. The inspection planning problem is formulated as a constrained open-loop optimal control problem and is solved using mixed integer programming (MIP) optimization. Quantitative and qualitative evaluation demonstrates the effectiveness of the proposed approach. 
\end{abstract}

\section{Introduction} \label{sec:Introduction}
Recent technological and scientific advancements in aerospace, avionics and artificial intelligence, in  conjunction with the cost reduction of electronic parts and equipment, have made increasingly popular the use of unmanned aerial vehicles (UAVs) in various application domains. UAVs have found widespread utilization in various tasks including emergency response and search-and-rescue missions \cite{PapaioannouTMC,Papaioannou2019,Papaioannou2020,Papaioannou2021a,Papaioannou2021b}, precision agriculture \cite{Tsouros2019,Radoglou2020}, wildlife monitoring \cite{Hodgson2016,Gonzalez2016}, and security \cite{PapaioannouJ1,PapaioannouJ2}.

One of the most important functionalities that will further enable the utilization of fully autonomous UAVs in the application domains discussed above is that of trajectory planning \cite{Gasparetto2015}, also known in the literature as motion/path planning \cite{Goerzen2010}. This technology is of crucial importance  in designing and executing automated UAV-based flight missions (i.e., automated UAV guidance, navigation and control), according to the requirements of the task at hand. In trajectory planning we are interested in delivering a collision-free motion between an initial and a final location within a given environment. Moreover, for many tasks including UAV-based automated maintenance operations, target search and infrastructure inspection, there is the need for finding the optimal trajectory which allows an autonomous UAV to utilize its sensors in order to cover or inspect every point within a given area or structure of interest. This problem is known in the literature as inspection planning (IP) or coverage path planning (CPP) \cite{Galceran2013}, and is the focus of this work. More specifically, during an automated inspection mission the UAV agent must autonomously plan its inspection trajectory which allows the efficient coverage/inspection of all points of interest on a given structure, while satisfying the vehicle's dynamic and sensing constraints.

Although a plethora of inspection planning approaches have been proposed in the literature, the technology has not yet reached the required level of maturity to fully support autonomous UAV operations. Towards this direction, in this paper we investigate the UAV-based inspection planning problem in 3D environments for cuboid-like structures (or objects) of interest, such as buildings, that need to be fully inspected. More specifically, we assume that a finite number of feature-points are scattered throughout the surface area of the structure of interest and must be inspected by an autonomous UAV agent. The UAV agent evolves in 3D space according to its dynamical model and is equipped with a camera sensor which exhibits a dynamic sensing range i.e., the size of the camera's projected field-of-view (FOV) on a given surface is a function of the distance between the UAV and that surface. Based on these assumptions, we propose a receding horizon mixed integer programming (MIP) inspection planning controller, for optimally determining the UAV's motion control inputs within a finite rolling planning horizon, subject to the UAV's sensing capabilities. The UAV's inspection trajectory is generated on-line by solving at each time-step a constrained open-loop optimal control problem until all feature-points are inspected (i.e., all feature-points are viewed through the UAV's camera).  Specifically, the contributions of this work are the following:

\begin{itemize}
    \item We propose a receding horizon inspection planning control approach which allows an autonomous UAV agent, governed by dynamical and sensing constraints, to inspect in 3D environments cuboid-like structures of interest (e.g., buildings).
    \item The inspection planning problem is formulated as a constrained optimal control problem and solved over a finite rolling planning horizon using mixed integer programming (MIP) optimization, allowing the on-line generation of the UAV's optimal inspection trajectory.
    \item Qualitative and quantitative evaluation demonstrates the performance of the proposed approach.
\end{itemize}

The rest of the paper is organized as follows. Section~\ref{sec:Related_Work} summarizes the related work on inspection planning with ground and aerial vehicles. Section \ref{sec:Preliminaries} discusses our assumptions and develops the system model, and Section \ref{sec:problem} formulates the problem tackled in this work. Then, Section \ref{sec:approach} discusses the details of the proposed inspection planning control approach and Section \ref{sec:Evaluation} evaluates the proposed approach. Finally, Section \ref{sec:conclusion} concludes the paper and discusses future work.

\section{Related Work}\label{sec:Related_Work}
Several approaches and algorithms can be found in the literature for the problem of autonomous inspection/coverage planning with single and multiple robots. In this section we give a brief overview of the most relevant techniques. A detailed survey regarding the various inspection/coverage planning techniques in the literature can be found in \cite{Galceran2013,Cabreira2019}.
Most notably, the problem of inspection/coverage planning with ground robots was investigated in \cite{Choset1998} and \cite{Danner2000}. Specifically, the authors in \cite{Choset1998}, propose a boustrophedon cellular decomposition coverage algorithm, in which the free-space of a 2D planar environment is a) first decomposed into non-intersecting regions or cells and b) the cells are then visited by the robot sequentially and covered with simple back-and-forth motions. The work in \cite{Danner2000} proposes a two-stage approach for inspecting polygonal objects, with the main objective being the computation of a path such that each point on the object's boundary is observed by the robot. The algorithm in \cite{Danner2000} first finds a set of sensing locations which allow full inspection of the polygonal object. In the second stage, the sensing locations found during the previous stage, are connected with the shortest path to generate the robot's inspection path.
Several other works \cite{Gabriely2001,Acar2002} have also proposed the decomposition of the free space into several non-overlapping regions, which can be individually covered and inspected with sweeping motions. Extensions \cite{Huang2001,Mnnadiar2010} of these approaches have investigated the optimal region traversal order and the optimal sweeping direction. 

The problem of 2D coverage planning was also investigated in the context of camera networks \cite{Mavrinac2013,Munishwar2013}, with the main objective being the optimal control and placement of cameras for full visual coverage of the monitoring space. The majority of the related work discussed so far, transforms the inspection/coverage planning problem to a path planning problem by firstly decomposing the area/object of interest into a number of non-overlapping cells which are then connected together with a path-finding algorithm to form the robot's path. Moreover, these approaches have mainly been tested in 2D environments and they do not consider the robot's dynamic and sensing behavior, i.e., they do not find the robot's control inputs which generate the inspection trajectory.

More related to the proposed approach is the work in \cite{Yi2006} which proposes a coverage control approach for guiding a mobile robot to completely cover a bounded two-dimensional region. In \cite{Yi2006} the 2D bounded surveillance environment is first covered with a minimum number of disks which exhibit a radius equal to the robots sensing range and then a neural network is used to plan the robot's coverage path. At a second stage, the generated path is adapted to the robot's kinematic constraints. Moreover, in \cite{Franco2015}, an energy optimized graph-based planner is proposed for UAV-based coverage in 2D discrete environments. Similarly, in \cite{Xu2011} a UAV-based terrain coverage approach is proposed for computing a trajectory through a known environment with obstacles that ensures coverage of the terrain while minimizing path repetition. In \cite{Wang2017} the terrain coverage problem with a UAV is investigated more realistically with the inclusion of photogrammetric constraints. The approach in \cite{Maza2007} tackle the terrain-coverage problem for rectilinear polygonal environments with multiple UAV agents. In this case, the environment is partitioned into multiple sub-regions, which are assigned to the UAVs according to their coverage capabilities. 

An off-line sampling-based 3D coverage planning approach for an underwater inspection robot is proposed in \cite{Englot2013}. Specifically, the authors propose a redundant roadmap algorithm and a watchman route algorithm \cite{watchman}, to allow the construction of a discrete set of stationary robot views which allow full coverage of the object of interest. The generated view configurations are then connected by solving an instance of the traveling salesman problem (TSP), although the generated path might be infeasible for robots with dynamical constraints. 

Finally, inspection planning approaches based on the next best view (NBV)/view-planning techniques \cite{Gonzalez2000,Dornhege2013,Jing2016} are concerned with the computation of sequence of viewpoints which results in complete scene coverage. These methods are usually applied in unknown environments and often provide only suboptimal results since they must solve instances of the set cover problem \cite{Zhang2009} and the traveling salesman problem (TSP) \cite{Laporte1996}. Moreover, these approaches focus on the selection of discrete sensor views rather than the construction of a continuous trajectory, which also accounts for the robots dynamical and/or sensing constraints. 


\section{Preliminaries} \label{sec:Preliminaries}

\subsection{Agent Dynamical Model} \label{ssec:agent_dynamics}
In this work an autonomous UAV agent maneuvers inside a bounded surveillance region $\mathcal{W} \subset \mathbb{R}^3$, with dynamics governed according to the following discrete-time dynamical model:
\begin{equation} \label{eq:agent_dynamics}
    x_{t+1} = A x_{t} + B u_{t}
\end{equation}
where $x_t = [p_t,\nu_t]\in \mathbb{R}^{6,1}$ is the state of the agent at time-step $t$, which consists of position $p_t  \in \mathbb{R}^{3,1}$ and velocity $\nu_t  \in \mathbb{R}^{3,1}$ components in 3D cartesian coordinates. The vector $u_t \in \mathbb{R}^{3,1}$ denotes the input control force applied in each dimension, which allows the agent to change its direction and speed. The matrices $A$ and $B$ are given by:
\begin{equation}
A = 
\begin{bmatrix}
    \text{I}_{3\times3} & \delta t~ \text{I}_{3\times3}\\
    \text{0}_{3\times3} & \alpha ~ \text{I}_{3\times3}
   \end{bmatrix},~
B = 
\begin{bmatrix}
    \text{0}_{3\times3} \\
     \beta  ~\text{I}_{3\times3}
   \end{bmatrix}
\end{equation}

\noindent where $\delta t$ is the sampling interval, $\text{I}_{3\times3}$ and $\text{0}_{3\times3}$ are the identity matrix and zero matrix respectively, both of dimension $3 \times 3$, and the parameters $\alpha$ and $\beta$ are given by $\alpha =  (1-\eta)$ and $\beta = \frac{\delta t}{m}$. Moreover, the parameter $\eta$ is used to model the air resistance and $m$ denotes the agent's mass.

%
%

\subsection{Agent Camera Model} \label{ssec:sensing_model}
\begin{figure}
	\centering
	\includegraphics[width=\columnwidth]{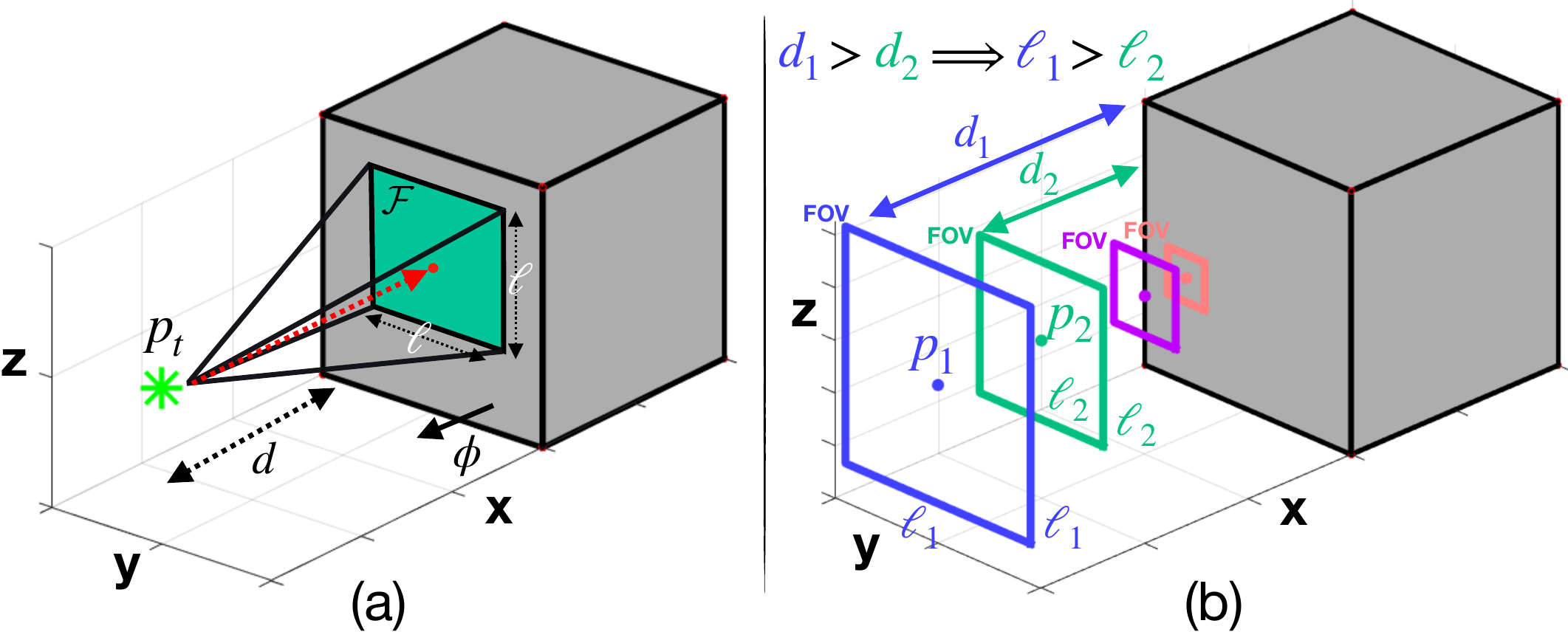}
	\caption{The figure illustrates the UAV's sensing model. (a) The UAV is equipped with a camera that has a view which is represented by a regular square pyramid whose apex (i.e., $p_t$) is directly above the centroid of its square base, and whose sides are 4 triangular faces meeting at the apex. $\mathcal{F}$ denotes the camera's projection, on the surface of the cuboid-like structure, which is represented by a square with length $\ell$. (b) The size of the projected FOV on a planar surface is a function of the distance $d$ between the agent position $p_t$ and the cuboid-like structure.}	
	\label{fig:fig1}
	\vspace{-5mm}
\end{figure}

The UAV agent is equipped with an onboard forward facing camera which is used for inspecting cuboid-like structures. Without loss of generality, we assume that the camera's horizontal and vertical field-of-view (FOV) angles are equal, and thus the projection of the camera's FOV $(\mathcal{F})$ on a planar surface has a square footprint with side length $\ell$. This is depicted in Fig. \ref{fig:fig1}(a), where the camera view is represented by a regular square pyramid whose apex (i.e., the center of the camera located at the lens) is directly above the centroid of its square base, and whose sides are 4 triangular faces meeting at the apex. We assume that the size of the projected FOV footprint is a linear function of the distance between the agent and the object of interest and thus:
\begin{equation}\label{eq:camera_model}
  \ell = g(d) = z_1 d + z_0
\end{equation}
where $d$ denotes the distance between the location of the agent and the structure that needs to be inspected, and the pair $(z_0,z_1)$ are the model's parameters. This is shown in Fig. \ref{fig:fig1}(b), where the size of projected square camera footprint decreases as the agent approaches the structure (i.e., the distance between the agent and the structure decreases and so does the size of camera FOV). Moreover, in this work it is assumed that the camera principal axis (depicted with the red line from the camera center perpendicular to the image plane in Fig. \ref{fig:fig1}(a)) is always parallel with the outward normal vector ($\phi$) of the plane which contains the face that is being viewed. In other words it is assumed that the agent automatically adjusts the onboard camera so that the viewing direction is parallel with the normal vector $\phi$ at all times.

\subsection{Structure to be inspected} \label{ssec:cuboids}
The 3D structure (or object) to be inspected by the UAV agent is represented in this work by a rectangular cuboid $\mathcal{C}$ of arbitrary size. A rectangular cuboid is convex polyhedron which exhibits six faces $f_i, i=[1,..,6]$, where opposite faces are equal and parallel, and each pair of adjacent faces meets in a right angle. The equation of the plane which contains the $i_\text{th}$ face of the cuboid is given by:

\begin{equation}\label{eq:plane_eq}
  \phi_i^\top \cdot x =  \gamma_i
\end{equation}

\noindent where $\phi_i^\top \cdot x$ denotes the dot product between $\phi_i^\top$ and $x$, $\phi_i \in \mathbb{R}^{3,1}$ is the outward normal vector to the plane which 
contains the $i_\text{th}$ face, $\gamma_i \in \mathbb{R}$ is a constant derived by the dot product of $\phi_i$ with a known point on the plane, and $x \in \mathbb{R}^{3,1}$ is an arbitrary point in 3D space. Thus, all points $x$ which satisfy the equality in Eqn. \eqref{eq:plane_eq} belong to the plane which contains the $i_\text{th}$ face of the cuboid. Consequently, a point $x \in \mathbb{R}^{3,1}$ resides inside the cuboid $\mathcal{C}$ when it satisfies all six inequalities:

\begin{equation}\label{eq:inside_cuboid}
  x \in \mathcal{C} \iff \phi_i^\top \cdot x \le \gamma_i,~ \forall i=[1,..,6]
\end{equation}

\noindent Equivalently, Eqn. \eqref{eq:inside_cuboid} can be written more compactly in matrix form as $\Phi x \le \Gamma$, where $\Phi$ is a matrix of dimensions $6 \times 3$, where the $i_\text{th}$ row corresponds to $\phi_i$ and $\Gamma$ is a column vector of dimensions $6 \times 1$, where the $i_\text{th}$ element corresponds to $\gamma_i$. 

The goal of the UAV agent is to inspect all faces of cuboid $\mathcal{C}$. More specifically, we consider the existence of a finite number of feature-points scattered on the cuboid's surface that must be viewed through the agent's camera in order for $\mathcal{C}$ to be visually inspected by the UAV agent. Let $\xi_j^i \in \mathbb{R}^{3,1}$ to represent the $j_\text{th}$ feature-point on the $i_\text{th}$ face of the cuboid $\mathcal{C}$. We say that cuboid $\mathcal{C}$ has been visually inspected by the UAV agent \textit{iff}:
\begin{equation}
   \xi_j^i \in \bigcup_{t \le \mathcal{T}_\text{max} ~\wedge~ i \in L} \mathcal{F}^i_{t}, ~ \forall j
\end{equation}

\noindent where $\mathcal{T}_\text{max}$ denotes the total mission inspection time, $\mathcal{F}^i_t$ denotes the projected camera FOV on the $i_\text{th}$ face of the cuboid at time-step $t$, and finally $L$ denotes the set of cuboid faces which contain feature-points that need to be inspected. In this work we assume that the three-dimensional map of the environment is readily available for the UAV agent to use during its mission, which was obtained through a 3D mapping procedure \cite{Xiao2013} executed prior to the inspection mission. During this 3D mapping procedure, the structure to be inspected has been reconstructed as a rectangular cuboid $\mathcal{C}$, and a fixed number of feature-points $\xi_j^i \in \mathbb{R}^{3,1}$ have been identified and extracted from its 3D point cloud. The UAV agent uses this 3D map to acquire the number of feature-points along with their location in order to plan its inspection mission. 


\begin{figure}
	\centering
	\includegraphics[scale=0.33]{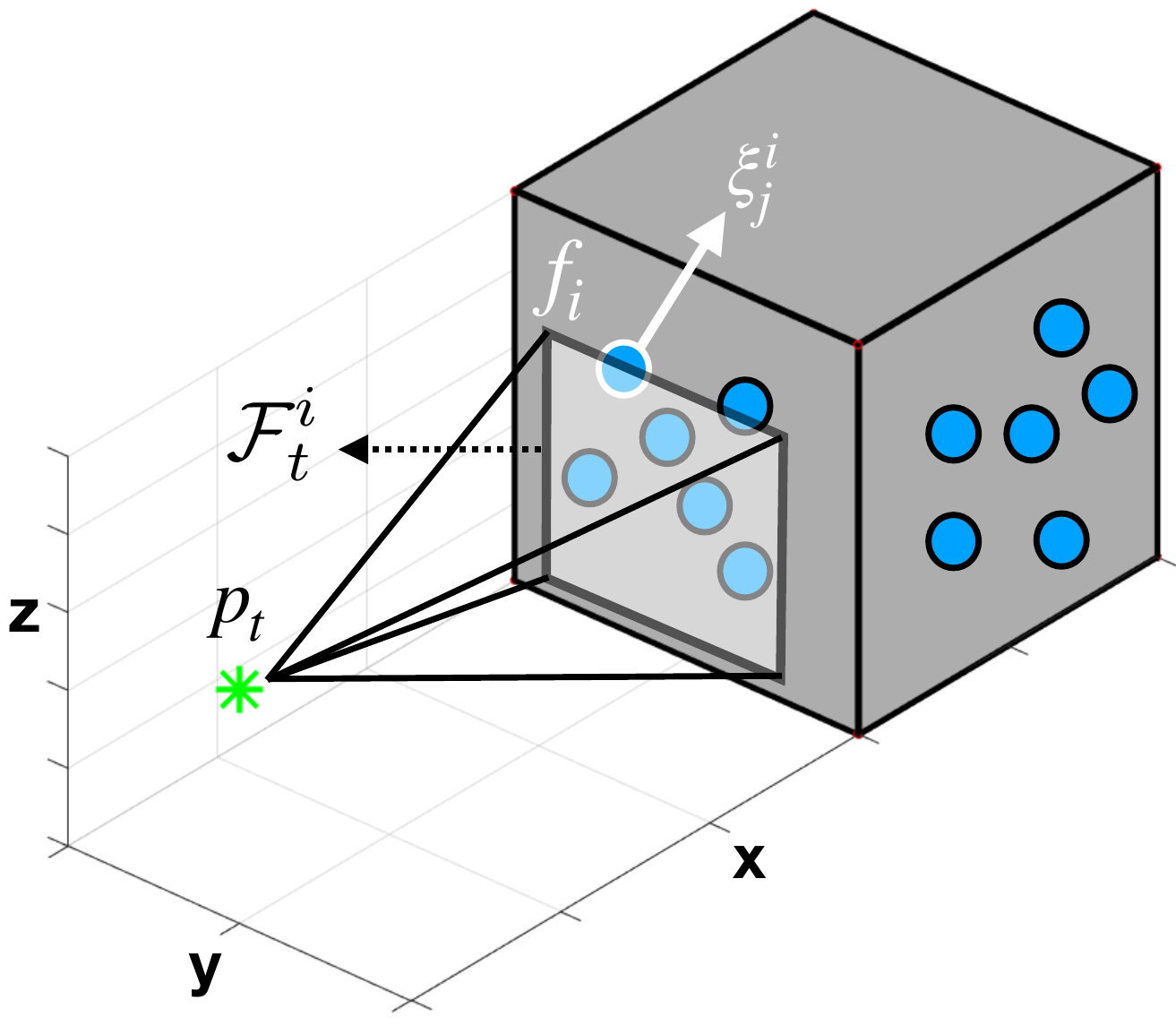}
	\caption{The UAV's objective is to inspect all $N_i$ feature-points $\xi^i_j, j=[1,..,N_i]$ scattered on face $f_i, i \in L$, for all $|L|$ faces which contain feature-points. The feature-point $\xi^i_j$ is inspected at time $t$ if it resides within the camera's projected FOV i.e., $\xi^i_j \in \mathcal{F}^i_t$ as illustrated in this figure.}	
	\label{fig:fig2}
	\vspace{-5mm}
\end{figure}

\section{Problem Statement}\label{sec:problem}
Let an autonomous UAV agent with initial state $x_t$ at time-step $t$, to evolve inside a bounded surveillance area $\mathcal{W} \subset \mathbb{R}^3$. Let a cuboid-like structure $\mathcal{C} \in \mathcal{W}$ to contain a number of feature-points $\xi^i_j, ~i \in L,~ j=[1,..,N_i]$, where $L$ denotes set of cuboid faces which contain feature-points and  $N_i$ denotes the total number of feature-points on the $i_\text{th}$ face of the cuboid. The inspection planning problem tackled in this work can be stated as follows: \textit{Given a sufficiently large mission inspection time $\mathcal{T}_\text{max}$, find the optimal UAV control inputs $U_t=\{u_{t|t},..,u_{t+T-1|t}\}, t~\le~\mathcal{T}_\text{max}$, over a rolling finite planning horizon of length $T$ time-steps, such that all feature-points on the cuboid's surface are visually inspected by the UAV agent at some point during the mission time i.e., $\xi^i_j \in \bigcup_{t \le \mathcal{T}_\text{max} ~\wedge~ i \in L}  \mathcal{F}^i_{t}, \forall j$}. In a high level form, the problem tackled in this work can be formulated as shown in Problem (P1).

\begin{algorithm}
\begin{subequations}
\begin{align}
&\textbf{Problem (P1):}~\texttt{High-level Controller} &  \nonumber\\
&~~~~~~~~~\underset{\mathbf{U}_t}{\arg \min} ~\mathcal{J}_\text{Inspection}, ~t \le \mathcal{T}_\text{max}& \label{eq:objective_P1} \\
&\textbf{subject to:} ~ \tau \in \{0,\ldots,T-1\} ~  &\nonumber\\
& x_{t+\tau+1|t} = A x_{t+\tau|t} + B u_{t+\tau|t} & \hspace{-10mm} \forall \tau \label{eq:P1_1}\\
& x_{t|t} = x_{t|t-1} & \label{eq:P1_2}\\
& \xi_j^i \in \bigcup_{t \le \mathcal{T}_\text{max} ~\wedge~ i \in L} \mathcal{F}^i_{t}, & \hspace{-30mm} \forall j=[1,..,N_i]\label{eq:P1_3}\\
& p_{t+\tau+1|t} \notin \mathcal{C} & \hspace{-30mm} \forall \tau \label{eq:P1_4}\\
& x_{t+\tau+1|t} \in \mathcal{X},~ u_{t+\tau|t} \in \mathcal{U}  & \hspace{-30mm} \forall \tau\label{eq:P1_5}
\end{align}
\end{subequations}
\vspace{-10mm}
\end{algorithm}

In this work the notation $x_{t^\prime|t}$ denotes the agent's future predicted state at time-step $t^\prime$ which is computed at time-step $t$. In essence we are looking to find the optimal future UAV control inputs $U_t=\{u_{t|t},..,u_{t+T-1|t}\}$,  which optimize the inspection objective function $\mathcal{J}_\text{Inspection}$ and which satisfy a certain set of constraints i.e., Eqn. \eqref{eq:P1_1} - Eqn. \eqref{eq:P1_4}. The constraints in Eqn. \eqref{eq:P1_1} and Eqn. \eqref{eq:P1_2} are due to the agent's dynamical model as discussed in Sec. \ref{ssec:agent_dynamics}, and the constraint in Eqn. \eqref{eq:P1_3} makes sure that all feature-points $\xi_j^i$ on the cuboid $\mathcal{C}$ will be viewed by the agent's camera at least once during the mission i.e., all feature-points must be included inside the agent's projected camera FOV. Then the collision avoidance constraint in Eqn. \eqref{eq:P1_4} makes sure that the UAV agent avoids collisions with the cuboid under inspection $\mathcal{C}$ at all times, and finally, the constraint in Eqn. \eqref{eq:P1_5} place the agent's state and control inputs within the desired operating bounds $\mathcal{X}$ and $\mathcal{U}$ respectively.

\section{UAV-based Receding Horizon Inspection Planning Control}\label{sec:approach}

Essentially, the inspection planning problem discussed in the previous section is posed in this work as a receding horizon constrained optimal control problem, in which the future optimal UAV control inputs $U_t=\{u_{t|t},..,u_{t+T-1|t}\}, t \le \mathcal{T}_\text{max} $ are computed at each time-step $t$, over a finite moving planning horizon of length $T$. The first control input of the sequence is then applied to the UAV and the problem is solved again for the next time-step. In the proposed approach the trajectory planning decisions are optimized based on the mission objective in an on-line fashion, and according to a set of mission constraints including a) the UAV's dynamical and sensing model, b) collision avoidance constraints and c) duplication of effort constraints. 


Next we discuss the details of the proposed receding horizon inspection planning controller for a single cuboid-like structure that needs to be inspected. Specifically, we have transformed the optimal control inspection planning problem shown in (P1), into a mixed integer quadratic program (MIQP), as shown in detail in problem (P2), which can be solved using readily available optimization solvers \cite{Anand2017}. We will begin the analysis of the proposed approach by first discussing the mission constraints i.e., Eqn. \eqref{eq:P2_1} - Eqn. \eqref{eq:P2_19}. Then, we discuss in detail how we have designed the multi-objective cost function i.e., Eqn. \eqref{eq:objective_P2} that drives the inspection planning mission.

\begin{algorithm}
\begin{subequations}
\begin{align}
&\textbf{Problem (P2):}~ \texttt{Inspection Controller} &  \nonumber\\
&~~~~~~~~~\underset{\mathbf{U}_t}{\arg \min} ~\mathcal{J}_\text{Inspection}, ~t \le \mathcal{T}_\text{max}& \label{eq:objective_P2} \\
&\textbf{subject to} ~ \tau \in \{0,\ldots,T-1\} \textbf{:}  &\nonumber\\
& x_{t+\tau+1|t} = A x_{t+\tau|t} + B u_{t+\tau|t} & \hspace{-40mm} \forall \tau \label{eq:P2_1}\\
& x_{t|t} = x_{t|t-1} & \label{eq:P2_2}\\
& d^\text{face}_{\tau,i} = |H_i p_{t+\tau+1|t} - C_i| &\hspace{-40mm} \forall \tau,i \label{eq:P2_4}\\
& \ell_{\tau,i} = g(d^\text{face}_{\tau,i}) &\hspace{-40mm} \forall \tau, i \label{eq:P2_5}\\
& J_{i,c} p_{t+\tau+1|t} + (M-K_{i,c})b^{1}_{\tau,i,c} \le M &\hspace{-40mm} \label{eq:P2_6} \\
&  &\hspace{-40mm} \forall \tau, i, c=[1,..,5] \notag \\
& 5b^2_{\tau,i} - \sum_{c=1}^5 b^1_{\tau,i,c} \le 0 &\hspace{-40mm} \forall \tau, i \label{eq:P2_7}\\
& \sum_{i=1}^L b^2_{\tau,i} \le 1 &\hspace{-40mm} \forall \tau \label{eq:P2_8}\\
& \Omega_{i,c} \xi^i_j b^{3}_{\tau,i,j,c} - \Omega_{i,c} p_{t+\tau+1|t} b^{3}_{\tau,i,j,c} - \frac{\ell_{\tau,i}}{2} \le 0 &\hspace{-40mm} \label{eq:P2_9} \\
&  &\hspace{-40mm} \forall \tau, i, j, c=[1,..,4] \notag\\
& 4b^4_{\tau,i,j} - \sum_{c=1}^4 b^3_{\tau,i,j,c} \le 0 &\hspace{-40mm} \forall \tau, i, j \label{eq:P2_10}\\
& \kappa^1_{\tau,i,j} = b^2_{\tau,i} \wedge b^4_{\tau,i,j} &\hspace{-40mm} \forall \tau, i, j \label{eq:P2_11}\\
& \kappa^2_{\tau,i,j} \le  \kappa^1_{\tau,i,j} + \mathcal{Q}_{i,j} &\hspace{-40mm} \forall \tau, i, j \label{eq:P2_12}\\
& \sum_{\tau=0}^{T-1} \kappa^2_{\tau,i,j} \leq 1&\hspace{-40mm} \forall i, j \label{eq:P2_12a}\\
& \kappa^2_{\tau,i,j} * d^\text{face}_{\tau,i} \le D_\text{max} &\hspace{-40mm} \forall \tau, i, j \label{eq:P2_13}\\
& \Phi_l p_{t+\tau+1|t} \geq \Gamma_l - M o_{\tau,l}& \hspace{-40mm} \forall \tau, l \label{eq:P2_14}\\
& \sum_{l=1}^6 o_{\tau,l} \leq 5 &\hspace{-40mm} \forall \tau \label{eq:P2_15}\\
& x_{t+\tau+1|t} \in \mathcal{X},~ u_{t+\tau|t} \in \mathcal{U}  & \hspace{-40mm} \forall \tau\label{eq:P2_16}\\
& b^{1}_{\tau,i,c}, b^{2}_{\tau,i}, b^{3}_{\tau,i,j,c}, b^4_{\tau,i,j} \in \{0, 1\}   & \hspace{-40mm} \forall \tau, i, j, c\label{eq:P2_17}\\
& \kappa^1_{\tau,i,j}, \kappa^2_{\tau,i,j}, \mathcal{Q}_{i,j}, o_{\tau,l} \in \{0, 1\}   & \hspace{-40mm} \forall \tau, i, j, l\label{eq:P2_18}\\
& i=[1,..,|L|], ~ j=[1,..,N_i], ~l=[1,..,6]  & \hspace{-40mm} \label{eq:P2_19}
\end{align}
\vspace{-10mm}
\end{subequations}
\end{algorithm}

\subsection{Inspection Constraints}

The first two constraints in Eqn. \eqref{eq:P2_1} and Eqn. \eqref{eq:P2_2} are due to the agent's dynamical model as already discussed in Sec \ref{ssec:agent_dynamics}. The next constraint shown in Eqn. \eqref{eq:P2_4} computes the distance between the agent's position $p_{t+\tau+1|t}$ (also abbreviated as $p_\tau$) and every face $f_i, i \in L$ of the cuboid-like structure $\mathcal{C}$ for all future time-steps $\tau \in \{0,\ldots,T-1\}$ inside the planning horizon, where for brevity we have used the notation $d^\text{face}_{\tau,i}$ to mean $d^\text{face}_{t+\tau+1|t,i}$. In Eqn. \eqref{eq:P2_4}, $H$ is a matrix with dimensions $|L|$-by-3, where $|L|$ denotes the number of cuboid faces which need to be inspected, and $C$ is a $|L|$-by-1 column vector. The distance $d^\text{face}_{\tau,i}$ between the agent with position $p_{\tau}$ and the $i_\text{th}$ face is defined here as the 1-norm distance between $p_{\tau}$ and its orthogonal projection on the plane which contains $f_i$, i.e., the perpendicular distance to the nearest point on the plane. For instance, let the plane (with equation $x=a$) which is parallel to the $zy$-axis to contain the $i_\text{th}$ face of the cuboid to be inspected, and the agent to be located in front of $f_i$ as shown in Fig. \ref{fig:fig1}, with $p_{\tau} = [p_\tau(x), p_\tau(y), p_\tau(z)]^\top$. In this example the $i_\text{th}$ row of the matrix $H$ is given by $H_i = [1, 0, 0]$ and $C_i = a$. Thus the distance between the agent and the face $f_i$ is computed as $d^\text{face}_{\tau,i} = |H_i p_{t+\tau+1|t} - C_i| = |p_\tau(x)-a|$. In a similar fashion, $H$ and $C$ are populated for all faces $|L|$ that need to be inspected and the distance between the agent and all the cuboid faces is computed for all time-steps inside the planning horizon using Eqn. \eqref{eq:P2_4}.

The next constraint shown in Eqn. \eqref{eq:P2_5} computes the size of the projected camera FOV on every face $i$ of the cuboid to be inspected for all time-steps $\tau$ inside the planning horizon. Specifically, the side length $\ell_{\tau,i}$ of the projected square camera FOV is computed as a function of the distance $d^\text{face}_{\tau,i}$ between the agent and each face $f_i$ i.e.:,

\begin{equation}
    \ell_{\tau,i} = g(d^\text{face}_{\tau,i}), ~~\forall \tau, i 
\end{equation}

\noindent where $g(.)$ is a linear or piecewise linear function with respect to the input (i.e., Eqn. \eqref{eq:camera_model}), and again $\ell_{t+\tau+1|t,i}$ has been abbreviated as $\ell_{\tau,i}$ for notational clarity.

\begin{figure}
	\centering
	\includegraphics[scale=0.35]{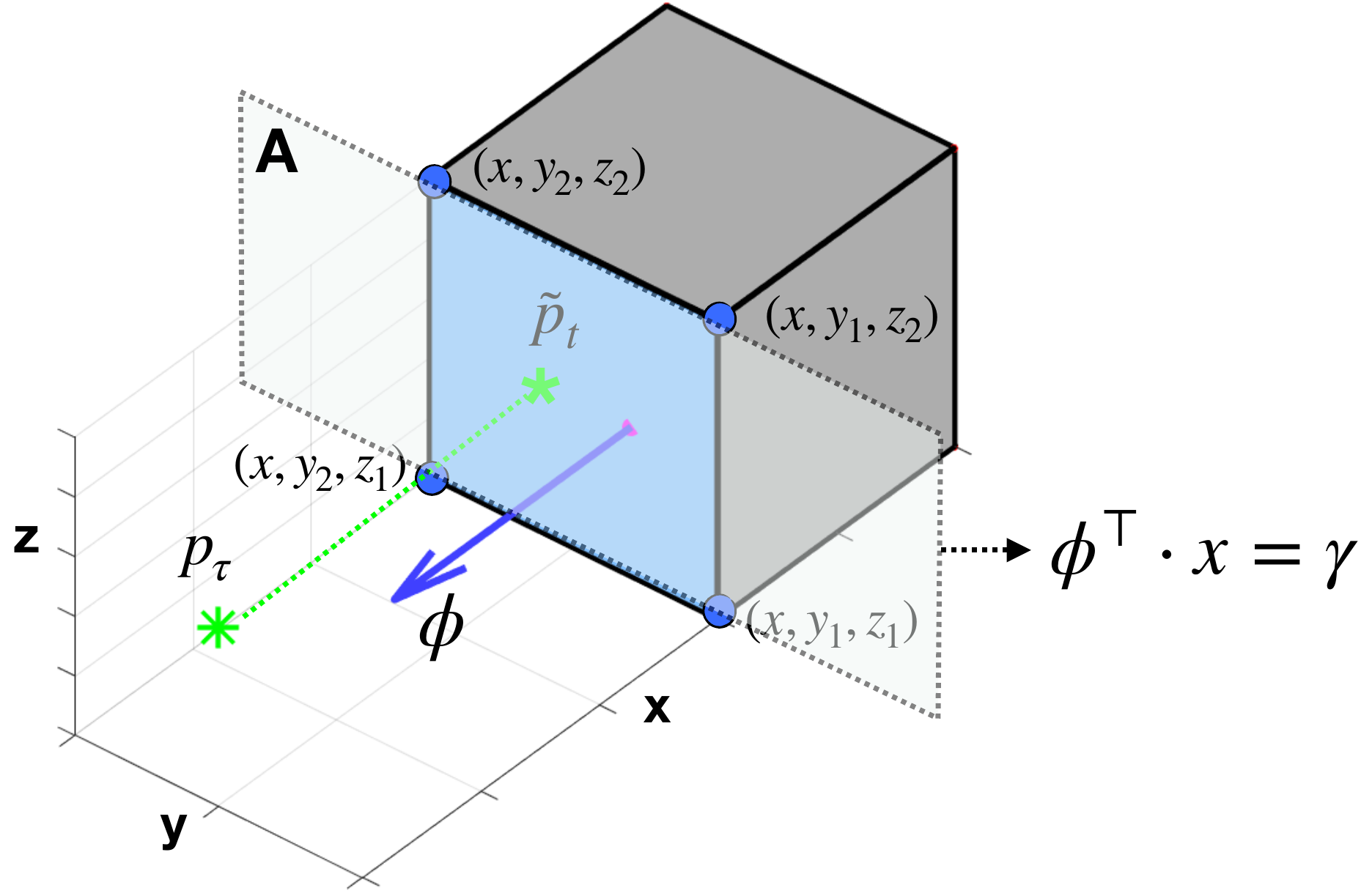}
	\caption{The UAV agent with position $p_\tau$ is viewing the blue face of the cuboid-like structure when a) the projection $\tilde{p}_\tau$ of its position on the that face resides inside the face's rectangular region (i.e., blue shaded area) and b) $p_\tau$ resides inside the positive half-space created by the plane A ($\phi^\top \cdot x = \gamma$) which contains the blue face as shown above, where $\phi$ denotes the outward normal vector on A.}	
	\label{fig:fig3}
	\vspace{-5mm}
\end{figure}

The constraints in Eqn. \eqref{eq:P2_6} - Eqn. \eqref{eq:P2_8} use the binary variables $b^1_{\tau,i,c}$ and $b^2_{\tau,i}$ to determine whether the agent's camera is viewing any of the cuboid's faces and if it does, identify on which face the camera's FOV is being projected. 
This is illustrated in Fig. \ref{fig:fig3}, where the agent with position $p_\tau$ is viewing the $i_\text{th}$ face of the cuboid (shown in blue color), with vertices $V = [v_1~  v_2~  v_3~  v_4]$ where $v_1 = [x, y_1, z_1]^\top$, $v_2 = [x, y_1, z_2]^\top$, $v_3 = [x, y_2, z_2]^\top$ and $v_4 = [x, y_2, z_1]^\top$. As shown in the figure, the face $f_i$ is contained within the plane $A$ with equation $\phi^\top \cdot x = \gamma$, where $\phi$ is the outward normal to the plane, $x$ is an arbitrary point on the plane and $\gamma$ is a constant. Let us denote the projection of $p_\tau$ on the plane $A$ as $\tilde{p}_\tau = [p_\tau(y), p_\tau(z)]^\top \in \mathbb{R}^{2,1}$. Now we can determine whether the agent is viewing face $f_i$ with the following two conditions which must be satisfied simultaneously: (a) the agent's projected position $\tilde{p}_\tau$ resides inside face $f_i$ i.e., $y_1 \le p_\tau(y) \le y_2$ and $z_1 \le p_\tau(z) \le z_2$ and (b) the agent must be located in front of face $f_i$ or equivalently the agent must reside inside the positive half-space formed by the plane $A$ which contains face $f_i$ i.e., $\phi^\top \cdot x \geq \gamma$ as illustrated in Fig. \ref{fig:fig3}. Condition (a) can be written more compactly in matrix form as $\tilde{J}_i \cdot p_\tau \le \tilde{K}_i$ where:

\begin{equation}
\tilde{J}_i =
    \begin{bmatrix}
       0 & -1 & 0 \\
       0 &  1 & 0 \\
       0 &  0 & -1 \\
       0 &  0 & 1 \\
     \end{bmatrix},~~ \text{and}~~
\tilde{K}_i =
     \begin{bmatrix}
       -y_1 \\
       y_2 \\
       -z_1 \\
       z_2 \\
     \end{bmatrix}
\end{equation}

\noindent Subsequently, the matrix $J_i$ and the column vector $K_i$ in Eqn. \eqref{eq:P2_6} are defined for face $f_i$ as $J_i = [\tilde{J}_i ~\phi^\top]$ and $K_i = [\tilde{K}_i ~\gamma]$. Thus $J_i$ and $K_i$ have dimensions $5 \times 3$ and $5 \times 1$ respectively. Thus, the agent is viewing face $f_i$ if all 5 constraints discussed above are being satisfied. The constraints in Eqn. \eqref{eq:P2_6} - Eqn. \eqref{eq:P2_8}, also shown below, can now be explained as follows:

\begin{align}
     & J_{i,c} p_{t+\tau+1|t} + (M-K_{i,c})b^{1}_{\tau,i,c} \le M, ~~\forall \tau, i, c=[1,..5] \notag\\
     & 5b^2_{\tau,i} - \sum_{c=1}^5 b^1_{\tau,i,c} \le 0, ~~\forall \tau, i \notag\\
     & \sum_{i=1}^L b^2_{\tau,i} \le 1, ~~\forall \tau \notag
\end{align}

First, the binary variable $b^1_{\tau,i,c}$ is used to indicate whether at time $\tau$ inside the planning horizon (or equivalently at time $t+\tau+1|t$), the constraint $c$ (out of 5) is true for face $f_i$. If the $c_\text{th}$ constraint is true then $b^1_{\tau,i,c}$ is activated i.e., $b^1_{\tau,i,c}=1$. Otherwise, $b^1_{\tau,i,c}=0$ and the inequality in Eqn. \eqref{eq:P2_6} becomes $J_{i,c} p_{t+\tau+1|t} \le M$ which is always true for a large constant $M$. Then, the constraint in Eqn. \eqref{eq:P2_7} uses the binary variable $b^2_{\tau,i}$ to determine whether $b^1_{\tau,i,c} = 1, \forall c=[1,..5]$, in which case $b^2_{\tau,i}$ is activated i.e., $b^2_{\tau,i}=1$. Thus the binary variable $b^2_{\tau,i}$ is used to determine whether face $f_i$ is being viewed by the agent at some point in time $\tau$. Observe that, Eqn. \eqref{eq:P2_7} is also satisfied when $b^2_{\tau,i}=0$. Finally, the constraint in Eqn. \eqref{eq:P2_8} makes sure that at any point in time $\tau$ at most one face is being viewed by the agent, which is added here for numerical stability purposes.

\begin{figure}
	\centering
	\includegraphics[scale=0.3]{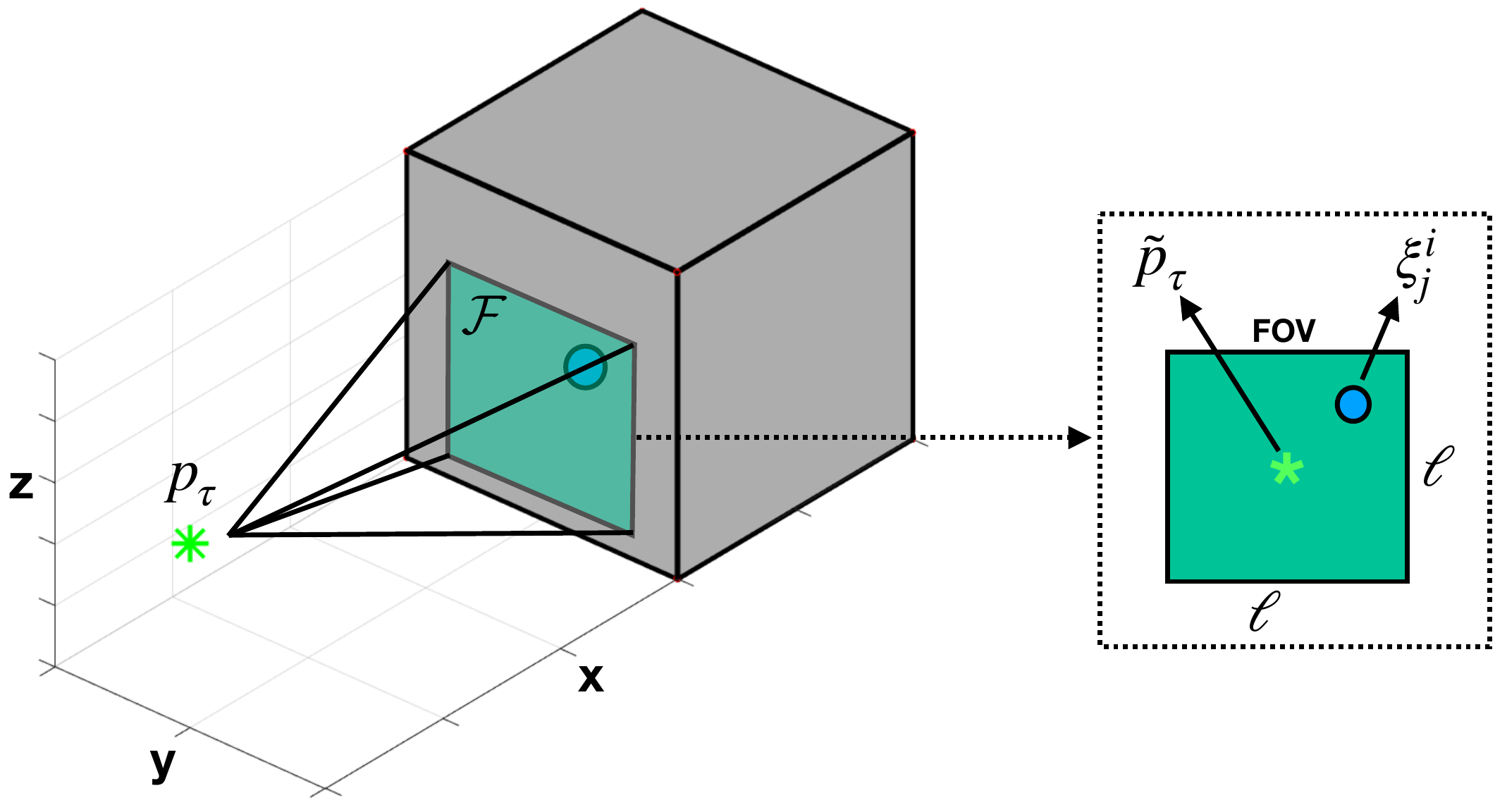}
	\caption{The 3D feature-point $\xi^i_j = [\xi^i_j(x), \xi^i_j(y), \xi^i_j(z)]^\top$ is planned to be inspected by the UAV agent $p_\tau$ at time-step $\tau$, when it resides inside the agent's camera projected FOV at time $\tau$ i.e., $\xi^i_j \in \mathcal{F}^i_\tau$. In the example illustrated above this is equivalent to the following constraints: $p_\tau(y)-\frac{\ell}{2} \le \xi^i_j(y) \le p_\tau(y)+\frac{\ell}{2}$ and $p_\tau(z)-\frac{\ell}{2} \le \xi^i_j(z) \le p_\tau(z)+\frac{\ell}{2}$. }	
	\label{fig:fig4}
	\vspace{-7mm}
\end{figure}

Moving forward with the analysis of the proposed receding horizon 3D inspection controller, the constraints in Eqn. \eqref{eq:P2_9} - Eqn. \eqref{eq:P2_10} (also shown below), use the binary variables $b^{3}_{\tau,i,j,c}$ and $b^4_{\tau,i,j}$ to determine whether the $j_\text{th}$ feature-point, on the $i_\text{th}$ face of the cuboid resides inside the agent's camera FOV projection $\tilde{\mathcal{F}}^i_\tau$ at time $\tau$.

\begin{align}
& \Omega_{i,c} \xi^i_j b^{3}_{\tau,i,j,c} - \Omega_{i,c} p_{t+\tau+1|t} b^{3}_{\tau,i,j,c} - \frac{\ell_{\tau,i}}{2} \le 0, ~~\forall \tau, i, j, c \notag\\
& 4b^4_{\tau,i,j} - \sum_{c=1}^4 b^3_{\tau,i,j,c} \le 0, ~~ \forall \tau, i, j \notag
\end{align}

\noindent The constraint in Eqn. \eqref{eq:P2_9} is illustrated more clearly with an example in Fig. \ref{fig:fig4}. First observe that the agent's position $p_\tau$ is determined by the constraints in Eqn. \eqref{eq:P2_1} and Eqn. \eqref{eq:P2_2}. The camera FOV projection $\tilde{\mathcal{F}}^i_\tau$ (colored green) on the $i_\text{th}$ face at time $\tau$ is a square, centered at $\tilde{p}_\tau = [p_\tau(y) p_\tau(z)]$ (i.e., the projection of the agent's position on the face $f_i$),  with side length $\ell_{\tau,i}$ given by Eqn. \eqref{eq:P2_5}, governed by the distance between the agent and face $f_i$ as determined by Eqn. \eqref{eq:P2_4}. The notation $\tilde{\mathcal{F}}^i_\tau, \forall i$ is used here to indicate the hypothetical FOV projection on the $i_\text{th}$ face of the cuboid at time $\tau$, assuming that the $i_\text{th}$ face is actually being observed. Thus, the procedure described in this paragraph computes the hypothetical FOV projection on all $|L|$ cuboid faces to determine which feature-points are included in each hypothetical FOV projection. As we discuss in the next paragraph at every time instance $\tau$ only one FOV projection is active which is instead denoted by $\mathcal{F}^i_\tau$, and determined as explained next.

To continue our discussion observe from Fig. \ref{fig:fig4}, that an arbitrary feature-point $\xi^i_j = [\xi^i_j(x), \xi^i_j(y), \xi^i_j(z)]^\top \in \mathbb{R}^{3,1}$ resides inside the hypothetical projected FOV $\tilde{\mathcal{F}}^i_\tau$ when:

\begin{subequations} \label{eq:f_in_F}
\begin{align} 
    & p_\tau(y)-\frac{\ell}{2} \le \xi^i_j(y) \le p_\tau(y)+\frac{\ell}{2}\\
    & p_\tau(z)-\frac{\ell}{2} \le \xi^i_j(z) \le p_\tau(z)+\frac{\ell}{2}
\end{align}
\end{subequations}

\noindent The constraints in Eqn. \eqref{eq:f_in_F} can be written in matrix form as:

\begin{equation}
     \begin{bmatrix}
       0 &  1 & 0 \\
       0 & -1 & 0 \\
       0 &  0 & 1 \\
       0 &  0 & -1 \\
     \end{bmatrix}
     \begin{bmatrix}
       \xi^i_j(x) \\
       \xi^i_j(y) \\
       \xi^i_j(z) \\
     \end{bmatrix}
     \leq
     \begin{bmatrix}
       0 &  1 & 0 \\
       0 & -1 & 0 \\
       0 &  0 & 1 \\
       0 &  0 & -1 \\
     \end{bmatrix}
     \begin{bmatrix}
       p_\tau(x) \\
       p_\tau(y) \\
       p_\tau(z) \\
     \end{bmatrix}
     + \frac{\ell_{\tau,i}}{2} \notag
\end{equation}

\noindent or more compactly as $\Omega_i \xi^i_j \leq \Omega_i p_\tau + 0.5\ell_{\tau,i}$. Therefore, $\Omega_i$ is a matrix with size 4-by-3 which encodes the 4 constraints that need to be satisfied in order for feature-point $\xi^i_j$ on the $i_\text{th}$ face of the cuboid to be included inside the agent's hypothetical FOV $\tilde{\mathcal{F}}_\tau^i$ at time $\tau$ when the agent is located at $p_\tau$. This is achieved with the binary variable $b^{3}_{\tau,i,j,c}$ which is activated when constraint $c$ (out of 4) is true. Otherwise $b^{3}_{\tau,i,j,c} = 0$. Subsequently, the binary variable $b^{4}_{\tau,i,j}$ in Eqn. \eqref{eq:P2_10} is activated when all 4 constraints are true i.e., $b^{3}_{\tau,i,j,c} = 1, \forall c=[1,..4]$. Thus, $b^{4}_{\tau,i,j}$ allows us to determine if at time $\tau$ the feature-point $j$ which is on the $i_\text{th}$ face of the cuboid, resides inside the hypothetical FOV projection $\tilde{\mathcal{F}}_\tau^i$.

Observe however, that with the binary variable $b^{4}_{\tau,i,j}$ alone we cannot determine whether feature-point $\xi^i_j$ is viewed by the agent. This is because $b^{4}_{\tau,i,j}$ does not encode which of the cuboid faces (if any) is the agent actually viewing at time $\tau$. Therefore, in order to actually determine if the feature-point $\xi^i_j$ is being observed by the UAV agent at time $\tau$ i.e., $\xi^i_j \in \mathcal{F}_\tau^i$, we use the constraint in Eqn. \eqref{eq:P2_11} as shown below:

\begin{equation}
 \kappa^1_{\tau,i,j} = b^2_{\tau,i} \wedge b^4_{\tau,i,j}, ~~ \forall \tau, i, j    \notag
\end{equation}

\noindent where we have combined the binary variables $b^2_{\tau,i}$ and $b^4_{\tau,i,j}$ with a logical conjunction. The resulting binary variable $\kappa^1_{\tau,i,j}$ is activated only when both $b^2_{\tau,i}$ and $b^4_{\tau,i,j}$ are true. Thus, the UAV agent views feature-point $\xi^i_j$ at time $\tau$, iff $\xi^i_j \in \tilde{\mathcal{F}}^i_\tau$ indicated by $b^4_{\tau,i,j}$ and the UAV agent views at the same time the $i_\text{th}$ face of the cuboid which is given by $b^2_{\tau,i}$.

As we have already mentioned, the UAV agent plans its inspection trajectory at every time-step $t$ over a moving finite planning horizon. Since in the majority of scenarios the inspection mission cannot be completed inside a single planning horizon, the agent needs to be equipped with some form of memory or record in order to keep track the mission progress (i.e., which feature-points are left to be inspected), and minimize the duplication of work (i.e., avoid inspecting feature-points that have already been inspected). To enable this functionality we use the constraint in Eqn. \eqref{eq:P2_12} i.e.,:

\begin{equation}
 \kappa^2_{\tau,i,j} \le  \kappa^1_{\tau,i,j} + \mathcal{Q}_{i,j}, ~~ \forall \tau, i, j \notag
\end{equation}

\noindent where the agent's memory is being realized with the 2D matrix $\mathcal{Q}_{i,j} \in \{0,1\}, ~i=[1,..|L|],~ j=[1,..,N_i]$. When the UAV agent inspects a particular feature-point $\xi^i_j$ the $(i,j)$-element of  $\mathcal{Q}_{i,j}$ is activated. Specifically, 

\begin{equation}
 \mathcal{Q}_{i,j} = 
  \begin{cases} 
   1, & \text{iff}~ \exists t \leq \mathcal{T}_\text{max} : \xi^i_j \in \mathcal{F}^i_t\\
   0, & \text{o.w} 
  \end{cases}
\end{equation}

\noindent where as we have previously mentioned $\mathcal{F}^i_t$ is the agent's projected FOV on the $i_\text{th}$ face of the cuboid at the current time-step $t$. Because $\kappa^2_{\tau,i,j}$ is also a binary variable, observe that with the constraint in Eqn. \eqref{eq:P2_12} the UAV agent has no incentive in inspecting (during the current and future planning horizons), a feature-point $\xi^i_j$ which has already been inspected i.e., $\mathcal{Q}_{i,j}=1$ (assuming Eqn. \eqref{eq:P2_12} is maximized). This allows the agent to minimize the duplication of work and plan inspection trajectories towards new feature-points. Next, the constraint in Eqn. \eqref{eq:P2_12a} discourages the agent from planning trajectories which inspect more than once the same feature-point inside the current planning horizon.

The quality of the visual inspection task of the cuboid-like structure inversely decreases with the distance between the agent and the feature-points to be inspected. In particular, we assume that for distances beyond a certain cut-off threshold, the amount of detailed captured by the agent's camera in not sufficient for reliably performing its inspection task. For this reason, we use the constraint in Eqn. \eqref{eq:P2_13} i.e.,

\begin{equation}
 \kappa^2_{\tau,i,j} * d^\text{face}_{\tau,i} \le D_\text{max}, ~~ \forall \tau, i, j 
\end{equation}

\noindent where $D_\text{max}$ denotes the cut-off distance beyond which the inspection task must not be performed. Consequently, the agent is forced to inspect feature-points $\xi^i_j$ at time $\tau$ (indicated by $\kappa^2_{\tau,i,j}$) i.e., $\xi^i_j \in \mathcal{F}^i_\tau$, from distances less than or equal to the cut-off distance.

Finally, the constraints in Eqn. \eqref{eq:P2_14} - Eqn. \eqref{eq:P2_15} implement collision avoidance constraints with the cuboid-like structure to be inspected. The objective here is to  make sure that the agent's position $p_\tau$ during the planning horizon does not resides inside the cuboid-like structure i.e., $p_\tau \notin \mathcal{C}$ which is accomplished as shown below:

\begin{align}
& \Phi_l p_{t+\tau+1|t} \geq \Gamma_l - M o_{\tau,l}, ~~ \forall \tau, l \notag\\
& \sum_{l=1}^6 o_{\tau,l} \leq 5, ~~ \forall \tau \notag
\end{align}

\noindent As we have already discussed in Sec. \ref{ssec:cuboids}, the agent is inside the cuboid-like structure at some time $\tau$ when $\Phi p_\tau~\leq~\Gamma$. Therefore, a collision can be avoided at time $\tau$ when $\exists~ l \in [1,..,6]: \phi^\top_{l} p_\tau \ge \gamma_l$, which is is implemented with the inequalities shown above. Specifically, we use the binary variable $o_{\tau,l}$ to determine whether the $l_\text{th}$ constraint (i.e., $\phi^\top_{l} p_\tau \ge \gamma_l$) is false, in which case $o_{\tau,l}$ is activated, i.e., $o_{\tau,l}=1$. Subsequently, the inequality in Eqn. \eqref{eq:P2_15} makes sure that the number of times $o_{\tau,l}$ is activated at a particular time $\tau$ is less than or equal to 5, which indicates that the agent is outside the cuboid-like structure at time $\tau$. The rest of the constraints in Eqn. \eqref{eq:P2_16} - Eqn. \eqref{eq:P2_19} restrict the agent's state and control inputs within the desired bounds and declare the various variables used.

\subsection{Inspection Objective} \label{ssec:inspection_obj}
To summarize, at each time-step $t \leq \mathcal{T}_\text{max}$ the autonomous UAV agent computes its future inspection trajectory $x_{t+\tau+1|t}, \forall \tau$ over a finite moving planning horizon i.e., $\tau \in \{0,..,T-1\}$ of length $T$, by taking into account the mission constraints in Eqn. \eqref{eq:P2_1} - Eqn. \eqref{eq:P2_19} and by optimizing the mission objective in Eqn. \eqref{eq:objective_P2}. Specifically, the agent's control inputs $U_t=\{u_{t|t},..,u_{t+T-1|t}\}$ are chosen such that the multi-objective cost function $\mathcal{J}_\text{Inspection}$ is minimized, which is defined in this work as follows:
\begin{equation}\label{eq:Jobj}
    \mathcal{J}_\text{Inspection} = \left(- \sum_{\tau=0}^{T-1}\sum_{i=1}^{|L|}\sum_{j=1}^{N_i}  \frac{\kappa^2_{\tau,i,j} (T-\tau)}{T} \right) +  w \times d^2_\text{target} 
\end{equation}


 \noindent The first term shown in Eqn. \eqref{eq:Jobj} is used here to maximize the number of unobserved feature-points over the planning horizon that reside inside the agent's projected FOV. As a reminder, the binary variable $\kappa^2_{\tau,i,j}$ indicates whether at time-step $t+\tau+1|t$, the feature-point $\xi^i_j$ is included inside the agent's projected FOV $\mathcal{F}^i_{t+\tau+1|t}$. Therefore, by minimizing the term inside the parenthesis, the UAV agent generates inspection plans which allow at each time-step $\tau$ of the planning horizon the maximum number of feature-points to be included inside the camera's projected FOV. Also note here that the weight factor $(T-\tau)T^{-1}$ which multiplies $\kappa^2_{\tau,i,j}$ drives the UAV agent to want to inspect feature-points at the earliest possible time. Moreover,  due to the constraint in Eqn. \eqref{eq:P2_12}, this sub-objective makes sure that the feature-points that are planned to be inspected have not been inspected in the past. Moreover, the constraint in Eqn. \eqref{eq:P2_13} makes sure that the inspection planning will take place within the camera's working distance i.e., all feature-points will be inspected at a distance less than or equal to the cut-off distance $D_\text{max}$.
 
\begin{figure}
	\centering
	\includegraphics[scale=0.3]{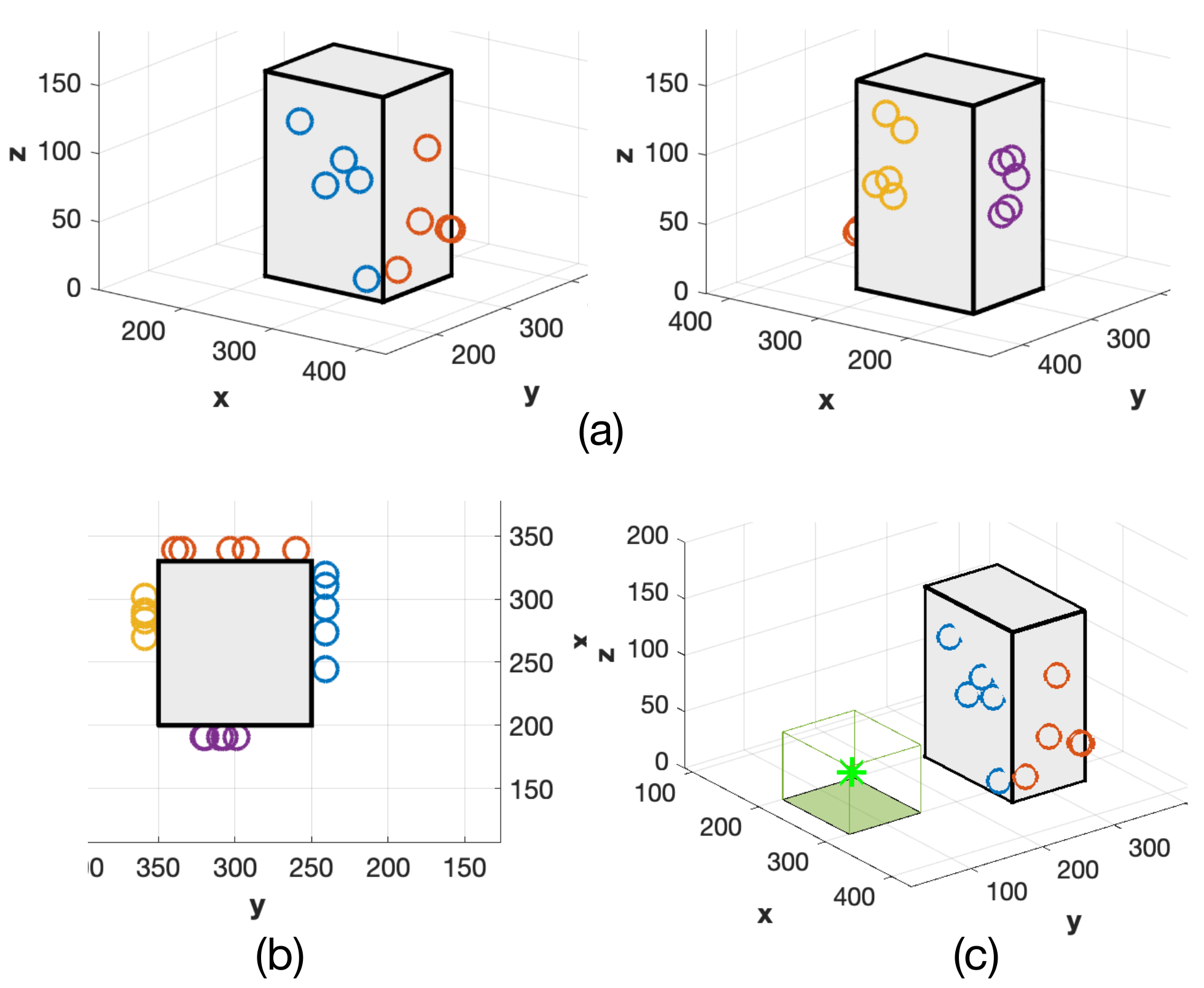}
	\caption{The figure illustrates an object of interest that needs to be inspected by the UAV agent. (a)(b) The feature-points that must be observed by the UAV agent are scattered on the object's surface, marked with circles. (c) The $\star$ denotes the agent's initial location.}	
	\label{fig:fig5}
	\vspace{-7mm}
\end{figure}

When the length of the planning horizon is not sufficiently large, solving the inspection planning problem by solely optimizing the term in the parenthesis becomes infeasible. This is because without a sufficiently large planning horizon the generated inspection trajectory would not be able to reach all feature-points for inspection, and the mission will fail. In order to make the problem independent of the length of the planning horizon and to increase the robustness of the proposed approach we include in the objective function an additional term (i.e., cost-to-go factor) as shown in Eqn. \eqref{eq:Jobj}, (where $d_\text{target} =\|p_{t+1|t} - \hat{\xi}^i_j\|_2$) which enables the agent to greedily move (inside each planning horizon) towards the nearest unobserved feature-point $\hat{\xi}^i_j$ i.e.,:

\begin{equation}\label{eq:next_target}
      \hat{\xi}^i_j = \underset{\xi^i_j \notin \mathcal{Q}_{i,j}}{\arg\min} \|p_{t|t} - \xi^i_j\|_2
\end{equation}

\begin{figure*}
	\centering
	\includegraphics[width=\textwidth]{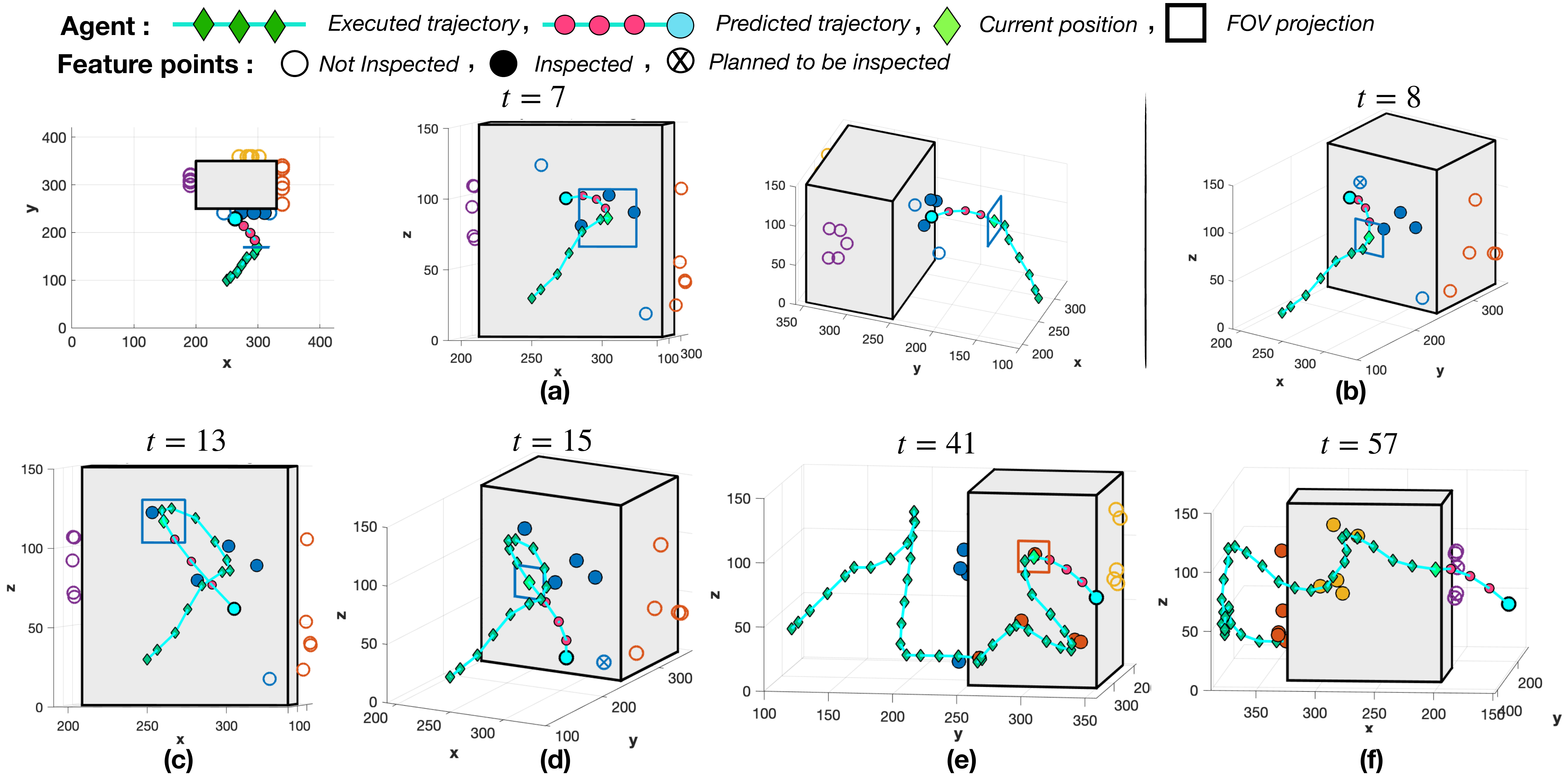}
	\caption{The figure illustrated the UAV's generated inspection trajectory for a simulated scenario with a cuboid-like object of interest containing 20 feature-points.}	
	\label{fig:fig6}
	\vspace{-7mm}
\end{figure*}

\noindent where the notation $\xi^i_j \notin \mathcal{Q}_{i,j}$ is used to denote feature-points that have not be observed and $p_{t|t}$ is the current UAV position. Thus Eqn. \eqref{eq:next_target} finds the nearest unobserved feature-point $\hat{\xi}^i_j$ with respect to the agent's current position and the minimization of the square of $d_\text{target}$ drives the agent's future position $p_{t+1|t}$ towards $\hat{\xi}^i_j$. Therefore, the agent can always move towards the unobserved feature-points even when no feature-points can be observed during the current planning horizon. Finally, $w$ is a tuning weight which determines the emphasis given to the two sub-objectives.

To summarize, we have presented a receding horizon inspection control approach which allows an autonomous UAV agent to inspect in 3D a cuboid-like structure. The proposed approach, based on the agent's dynamical and sensing model, generates inspection plans that enable the UAV agent to observe a finite number of feature-points scattered throughout the surface of a cuboid-like structure by appropriately selecting its control inputs inside a rolling planning horizon. The mission objective in Eqn. \eqref{eq:Jobj} aims at maximizing the number of unobserved future-points planned to be inspected inside the planning horizon and thus minimizing the mission time. Finally, the mission objective in Eqn. \eqref{eq:Jobj} also makes sure that a feasible solution can be obtained for planning horizons of arbitrary sizes. 

\section{Evaluation} \label{sec:Evaluation}

\subsection{Simulation Setup} \label{ssec:sim_setup}
For the evaluation of the proposed approach we have used the following setup: The UAV's dynamical model is according to Eqn. \eqref{eq:agent_dynamics} with $\delta t=1$s. The UAV's mass $m$ and the air resistance coefficient $\eta$ are set to $3.35$kg and respectively $0.2$. The UAV's control input $u_t$ is bounded in each dimension within the interval $[-20,20]$N, the UAV's acceleration $\nu_t$ is limited in each dimension within the interval $[-15,15]$ m/s and the parameters $z_0$ and $z_1$ are set to 10 and 0.5 respectively. We assume that the agent is maneuvering inside a bounded surveillance region of dimensions 500m by 500m by 250m,  and its objective is to inspect a cuboid-like object of interest  of dimensions 130m by 100m by 150m as shown in Fig. \ref{fig:fig5}. Specifically, the UAV agent needs to inspect all feature-points scattered on the object's surface (sampled uniformly) denoted by the colored circles (using different color per face) in  Fig. \ref{fig:fig5}(a)(b). As shown, each of the object's lateral faces contains 5 feature-points that need to be inspected by the agent. In this example the top and bottom face do not contain any feature-points, thus making the total number of feature-points that need to be inspected 20. Finally, the planning horizon $T$ is set to 5 time-steps, the maximum mission time $\mathcal{T}_\text{max}$ is set to 100 time-steps, the tuning weight $w$ in Eqn. \eqref{eq:Jobj} is set to $w=0.01$ and $D_\text{max}=100$m. Our evaluation has been conducted on a 2GHz laptop computer, with 8GB of RAM, running the Gurobi v9 solver.

\subsection{Results}
The UAV agent starts its mission from the $(x,y)$-coordinates $(250, 100)$, hovering 30m above the ground as shown in Fig. \ref{fig:fig5}(c). As depicted in Fig. \ref{fig:fig6}, the UAV's executed trajectory is denoted with $-\diamond-$ whereas the UAV's predicted trajectory (i.e., planned trajectory) is denoted by $-\circ-$. In Fig. \ref{fig:fig6}, the feature-points which have not yet been inspected are shown with clear circles, whereas feature-points which have been inspected by the UAV agent are shown in solid circles. Moreover, feature-points which have been planned to be inspected at some point in the future inside the planning horizon will be shown as $\otimes$. As shown in Fig. \ref{fig:fig6}, initially the UAV agent begins by approaching the face containing the blue feature-points. More specifically, between time-steps 1 and 6 the UAV agent is approaching the object of interest in order to place itself within the specified working distance $D_\text{max}$ according to the constraint in Eqn. \eqref{eq:P2_13} i.e., all feature-points must be inspected from a distance less or equal to $D_\text{max}$. 
At each time-step the UAV agent solves a finite-horizon optimal control problem which seeks to find the control inputs inside the planning horizon which will maximize the number of feature-points that can be inspected, as discussed in detail in Sec. \ref{ssec:inspection_obj}. This is shown in Fig. \ref{fig:fig6}(a), where at time-step $t=7$, the agent manages to inspect simultaneously 3 feature-points from a single location. As it is shown, the 3 feature-points which are marked with solid blue circles reside inside the UAV's projected FOV denoted with a blue square. At the subsequent time-step, also observe that the agent's prediction plan is headed towards the nearest unobserved feature-point according to the objective function defined in Eqn. \eqref{eq:Jobj}. As it is shown in Fig. \ref{fig:fig6}(b) the feature-point with $(x,y,z)$ coordinates $[245, 250, 120]$ has been marked for inspection as illustrated by the $\otimes$ symbol. Then, at time-step $t=13$ the same feature-point is inspected as shown in Fig. \ref{fig:fig6}(c). Observe that at $t=13$ this feature-point resides within the agent's projected FOV. Subsequently, the UAV agent moves towards the nearest unobserved feature-point as shown in Fig. \ref{fig:fig6}(d). The figure also illustrates the UAV's executed trajectory so far, along with its current prediction plan.
Once all blue feature-points have been inspected, the UAV agent moves to the next face as depicted in Fig. \ref{fig:fig6}(e). Here we can observe the generated UAV's trajectory for inspecting the orange feature-points (time-step $t=41$). Finally, at time-step $t=57$, the UAV agent has completed the inspection of the yellow feature-points as shown in Fig. \ref{fig:fig6}(f), and is moving towards the object's final face which contains the purple feature-points.

\begin{figure}
	\centering
	\includegraphics[width=\columnwidth]{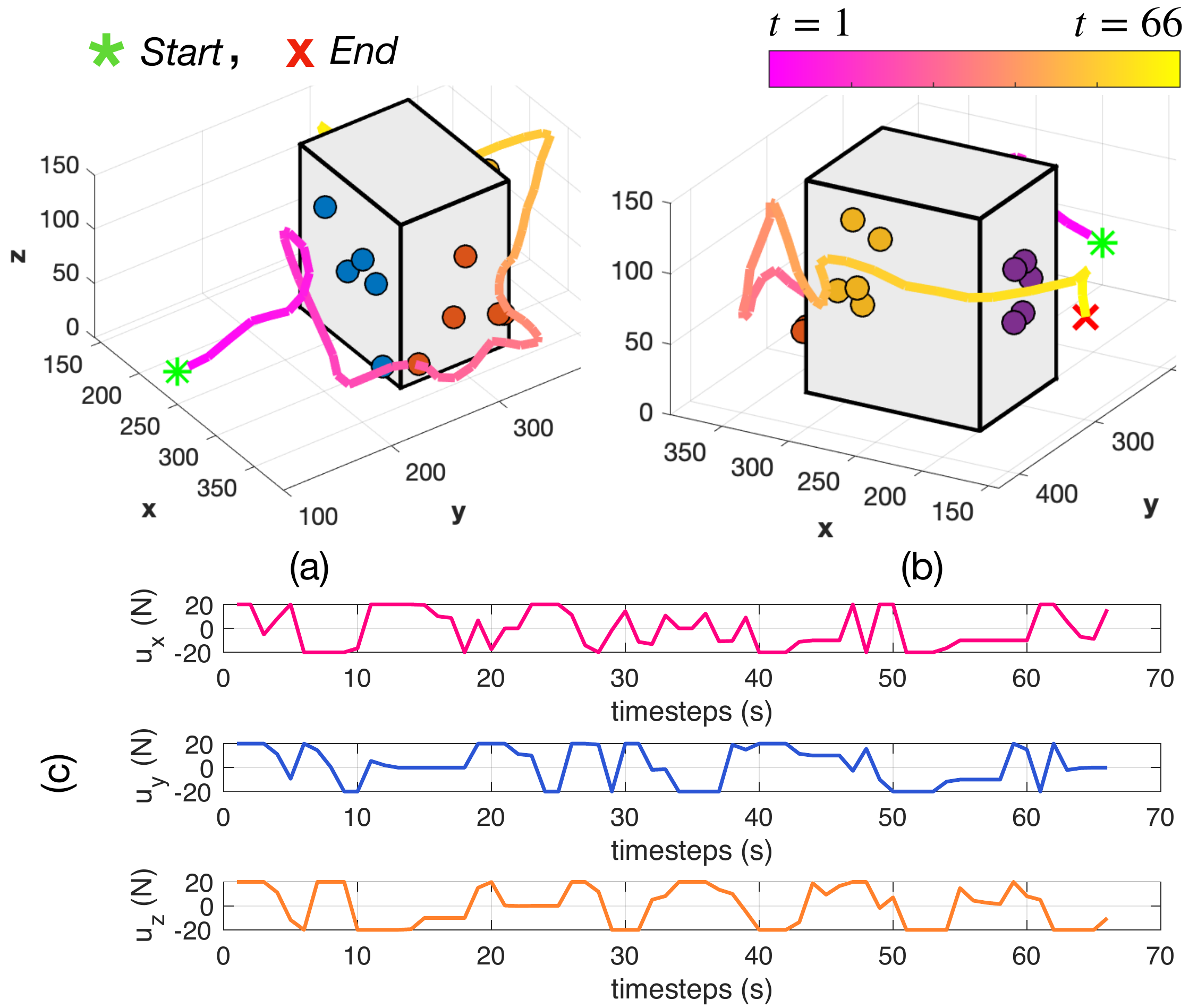}
	\caption{The figure illustrates: (a)(b) The UAV's inspection plan in 3D used to inspect 20 feature-points scattered on the surface of the object of interest. (c) The control inputs used for generating the inspection trajectory.}	
	\label{fig:fig8}
	\vspace{-7mm}
\end{figure}

\begin{figure}
	\centering
	\includegraphics[width=\columnwidth]{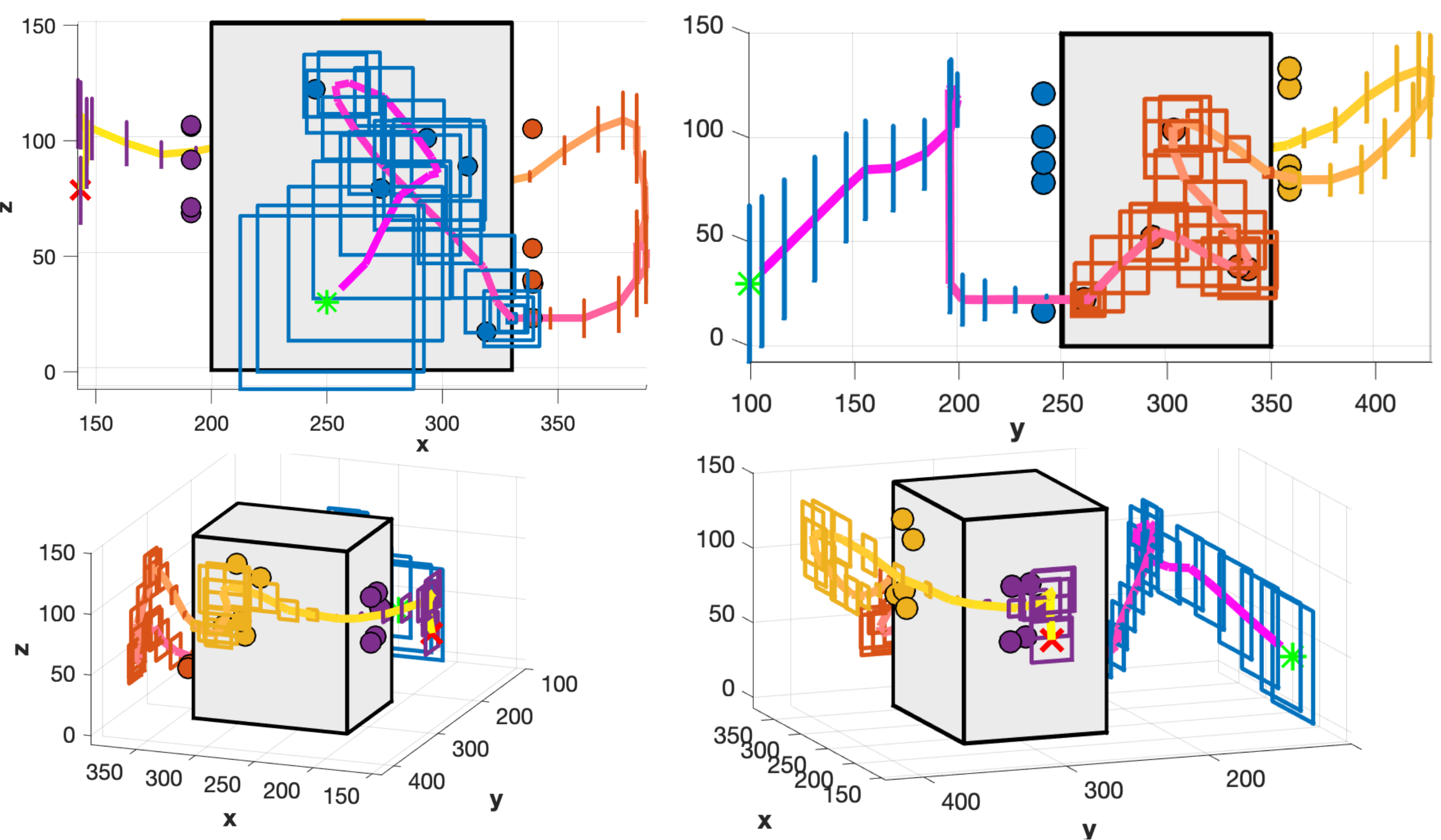}
	\caption{The figure illustrates the UAV's projected FOV along its inspection trajectory, and its size in terms of side length.}	
	\label{fig:fig7}
	\vspace{-1mm}
\end{figure}

Figure \ref{fig:fig8} shows the final inspection trajectory along with the applied control inputs. In this figure, the UAV's start and stop locations are marked with $\star$ and $\times$ respectively, and the inspection trajectory is color-coded according to the mission elapsed time. As it is shown in the figure all feature-points are inspected within 66 time-steps. Finally, Figure \ref{fig:fig7} shows the UAVs projected FOV along its inspection trajectory. Observe, that the UAV's control inputs are selected in such a way so that resulting FOV projections contain all the feature-points that need to be inspected, as indicated by the colored solid circles.

We conclude our evaluation by discussing the computational complexity of the proposed approach.
In general, the main optimization algorithm which is employed in order to tackle mixed integer programs (MIP) is that of branch-and-bound \cite{Mitchell2002}, and its complexity is usually due to the number of integral variables that are being utilized. Specifically, a branch-and-bound algorithm constructs a search tree by enumerating in systematic and consistent way candidate solutions for the MIP problem. Each node of this tree includes the original MIP problem constraints plus additional constraints on the bounds of the integer variables. The algorithm, proceeds by exploring nodes of the tree i.e., by solving a linear programming relaxation problem after dropping all integrality constraints. When the solution to the linear program consists of an integer constrained variable with a fractional value (i.e., $x=f$), the algorithm generates a new branch for this variable consisting of two sub-problems (i.e., nodes) where new integrality constraints are imposed (i.e., $x\le \lfloor{f}\rfloor$ and $x \ge \lceil{f}\rceil$). Therefore, the size of this search tree grows with the number of integral variables and as a consequence the computational complexity grows as well. 

As we have discussed in Sec. \ref{sec:approach} the proposed approach uses a number of binary variables in order to implement the desired inspection planning behavior. As shown by the constraints in Eqn. \eqref{eq:P2_17}-\eqref{eq:P2_18} the number of required binary variables depends mainly on the number of feature-points that need to be inspected and  on the length of the planning horizon. To better understand the computational complexity of the proposed controller, a Monte-Carlo simulation was conducted, where for the same object of interest we have varied the number of feature-points that need to be inspected and the length of the planning horizon, running 20 trials for each configuration and measuring the average runtime until the optimal solution is found. 
More specifically, we have run the proposed inspection planner with 12, 20 and 32 feature-points (in each trial the future points are randomly scattered on the object's surface area, with equal number of points per face), and with planning horizons of length 3, 8 and 15 time-steps. For this experiment the agent is always initialized from the same location, and the rest of the simulation parameters are set according Sec. \ref{ssec:sim_setup}. 

Table~\ref{tbl:tbl1} summarizes the results of this experiment in terms of the average execution time (i.e., the time required by the solver to find the optimal solution). In particular, Table~\ref{tbl:tbl1} shows the average time (taken over the $20$ trials) for each combination of the parameters. The results verify that the computational complexity increases as the number of feature-points and the length of the planning horizon increase. As it is shown, the length of the planning horizon has the largest impact on the performance of the proposed approach in terms of runtime (observe that the planning horizon is involved in every constraint listed in (P2)). Nevertheless, for some of the configurations of the parameters the proposed approach shows potential for real-time execution, considering that these results have been obtained on a 2GHz laptop computer. It is also worth noting that additional computational savings can be obtained through various heuristics and approximations \cite{Klotz2013,Naik2021,Hendel2021} which can provide adequate near-optimal MIP solutions in real-time. A more thorough investigation of the real-time performance of the proposed approach and its real-world implemetation will be investigated in future works.

\begin{table}\normalsize
\label{tbl:tbl1}
\begin{center}
  \begin{tabular}{|c|c|c|}
   \hline
    \multicolumn{3}{|c|}{Avg. Execution Time (sec)} \\
    \hline
    \# feature-points & Horizon Length & Runtime \\
    \hline\hline
    12 & 3  & 0.645  \\ \hline
    12 & 8  & 1.040  \\ \hline
    12 & 15  & 4.073  \\ \hline
    20 & 3 & 0.621   \\ \hline
    20 & 8 & 1.376  \\ \hline
    20 & 15 & 6.143  \\ \hline
    32 & 3 & 0.653   \\ \hline
    32 & 8 & 1.552  \\ \hline
    32 & 15 & 9.427  \\ \hline
  \end{tabular}
\end{center} 
\vspace{-8mm}
\end{table}

\section{Conclusion} \label{sec:conclusion}

In this work, we have tackled the problem of UAV-based automated planning, guidance and control for 3D inspection missions. We have formulated the inspection planning problem as a constrained receding horizon optimal control problem, in where the UAV's control inputs are optimally determined to enable the generation of efficient inspection trajectories of cuboid-like structures. We have derived a mixed-integer mathematical program which can be solved using off-the-shelf optimization solvers, and we have demonstrated its effectiveness through synthetic qualitative and quantitative experiments. In the future, we plan to extend the proposed approach to multiple UAV agents, and investigate its real-world performance.

\section*{Acknowledgments}

This work is supported by the European Union’s Horizon 2020 research and innovation programme under grant agreement No 739551 (KIOS CoE), and from the Republic of Cyprus through the Directorate General for European Programmes, Coordination and Development.

\flushbottom
\balance

\bibliographystyle{IEEEtran}
\bibliography{IEEEabrv,main} 

\end{document}